\documentclass[10pt,twocolumn,letterpaper]{article}

\usepackage{wacv}
\usepackage{times}
\usepackage{epsfig}
\usepackage{graphicx}
\usepackage{amsmath}
\usepackage{amssymb}

\makeatletter
\@namedef{ver@everyshi.sty}{}
\makeatother
\usepackage{booktabs}       
\usepackage{breqn}
\usepackage{float}
\usepackage{lipsum}
\usepackage{stfloats}
\usepackage{multirow}
\usepackage{pgfplots}
\usepackage{pgfplotstable}
\usepackage{tikz}
\usepackage{verbatim}
\pgfplotsset{width=7cm,compat=1.13}
\usepackage{wrapfig}

\usepackage[pagebackref=true,breaklinks=true,letterpaper=true,colorlinks,bookmarks=false]{hyperref}

\wacvfinalcopy 

\def\wacvPaperID{424} 

\begin{document}

\title{Stochastic Gradient Descent with \\ Hyperbolic-Tangent Decay on Classification}

\author{Bo-Yang Hsueh \\
Department of Computer Science\\ National Chiao Tung University \\
{\tt\small byshiue@gmail.com}
\and
Wei Li \\
Department of Computer Science\\ National Chiao Tung University \\
{\tt\small fm.bigballon@gmail.com}
\and
I-Chen Wu \\
Department of Computer Science\\ National Chiao Tung University \\
{\tt\small icwu@cs.nctu.edu.tw}
}

\maketitle
\ifwacvfinal\thispagestyle{empty}\fi

\begin{abstract}
   Learning rate scheduler has been a critical issue in the deep neural network training. Several schedulers and methods have been proposed, including step decay scheduler, adaptive method, cosine scheduler and cyclical scheduler. This paper proposes a new scheduling method, named hyperbolic-tangent decay (HTD). We run experiments on several benchmarks such as: ResNet, Wide ResNet and DenseNet for CIFAR-10 and CIFAR-100 datasets, LSTM for PAMAP2 dataset, ResNet on ImageNet and Fashion-MNIST datasets. In our experiments, HTD outperforms step decay and cosine scheduler in nearly all cases, while requiring less hyperparameters than step decay, and more flexible than cosine scheduler. Code is available at \url{https://github.com/BIGBALLON/HTD}. 
\end{abstract}

\section{Introduction}

Deep Neural Networks (DNNs) are currently the best-performing method for many classification problems. The variants of DNN have significant performance on many areas. For example, Convolutional Neural Network (CNN) \cite{LeNet} is widely used in image classification, object localization and detection. Recurrent Neural Network (RNN) \cite{LSTM} is widely used in language translation and natural language processing. 

Training DNN is usually considered as the non-convex optimization problem. Stochastic gradient descent (SGD) is one of the most used training algorithms for DNN. Although there are many different optimizers like Newton and Quasi-Newton methods \cite{OPT_book} in tradition, these methods are hard to implement and need to handle the problem of large cost on computing and storage. Compared to them, SGD is simpler and has good performance. The update direction of SGD is determined by the gradient of loss function. The parameters $\theta_t$ (weights) at time $t$ are updated by $\theta_t = \theta_{t-1} - \alpha_t \nabla_\theta L$, where $L$ is a loss function and $\alpha_t$ is the learning rate at time $t$. 

Unfortunately, it has been hard to tune the learning rate. A large learning rate makes the training diverge, while a small learning rate makes the training converge slowly. For a better performance, one usually needs to experiment with a variety of learning rates during the training process. A method of scheduling leaning rate, called a \textit{learning rate scheduler} in this paper, is used to change the learning rate during the training progress. 

There are many different schedulers used in the past, including step decay, adaptive learning rate methods \cite{AdaGrad, Adam, RMSProp, ADADELTA}, SGDR \cite{SGDR}, and so on. Step decay can get the ideal results in theory, but the process of tuning the learning rate is tedious and time-consuming. Adaptive methods can adjust the step size of each parameter by themselves, but the final performance are usually worse than fine-tuned step decay. SGDR used the cosine function to perform cyclic learning rates. Pure cosine scheduler (without cyclic) is a special case of SGDR, but performs better than step decay and cyclic cosine scheduler in about half of their experiments. However, users are only able to adjust the maximum and minimum learning rate of cosine scheduler, which is less flexible than step decay.

This paper proposes a new learning rate scheduler, called as Hyperbolic-Tangent decay (HTD) scheduler. Compared to step decay scheduler, HTD has less hyperparameters to tune and performs better than step decay scheduler in all experiments. Compared to cosine scheduler, HTD requires slightly more hyperparameters to tune for higher performance, and outperforms cosine schedulers in nearly all experiments in this paper. 

Section \ref{related_work} reviews some optimizers and learning rate schedulers proposed in the past. Section \ref{sec_HTD} describes the proposed HTD scheduler. Section \ref{experiments} shows experiment results of HTD against other learning rate schedulers on different architectures and datasets. Section \ref{conclusion} concludes the contributions of this paper and discusses the future work.

\section{Related work}
\label{related_work}

This section first reviews some optimizers and learning rate schedulers proposed in the past. Then, we review the traditional optimizers like stochastic gradient descent (SGD), SGD with momentum \cite{Momentum}, and Nesterov momentum \cite{Nesterov_theory, Nesterov_DNN}. Next, we review the adaptive learning optimizers including Adam, AdaDelta and so on \cite{AdaGrad, Adam, RMSProp, ADADELTA}. Finally, we review learning rate schedulers proposed last few years, including stochastic gradient descent with warm restarts (SGDR) \cite{SGDR}, cyclical learning rate (CLR) \cite{CLR} and exponential decay sine wave learning rate (ES-Learning) \cite{ESLearning}.

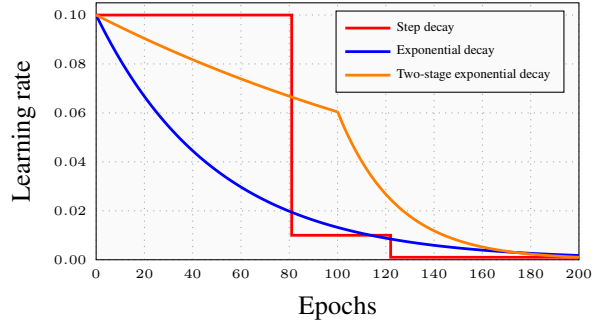
\begin{figure}
  \centering
  \begin{tikzpicture}
    \begin{axis}[
    xlabel={Epochs},
    ylabel={Learning rate},
    xtick={0,20,...,200},
    ytick={0,0.02,...,0.12},
    xmin=0,
    xmax=200,
    ymin=0,
    ymax=0.105,
    width=8cm,
    height=5cm,
    tick style={draw=none},
    y tick label style={
      /pgf/number format/.cd,
      fixed,
      fixed zerofill,
      precision=2,
      /tikz/.cd
    },    
    grid,
    grid style={dotted,gray!70},
    yticklabel style = {font=\tiny},
    xticklabel style = {font=\tiny},
    legend style={font=\tiny,legend pos=north east,cells={anchor=west},fill=gray!4},
    axis background/.style={fill=gray!4}
    ] 
    
    \addplot[
    color=red,
    line width=1.05pt
    ]
    coordinates {
    (0,0.1)(81,0.1)(81,0.01)(122,0.01)(122,0.001)(200,0.001)
    };
    \addlegendentry{Step decay}
    
    \addplot [
    domain=0:200, 
    samples=1000, 
    color=blue,
    line width=1.05pt
    ]
    {0.1*0.98^x};
	\addlegendentry{Exponential decay}
   
    \addplot[
    domain=0:100,
    samples=500,
    color=orange,
    line width=1.05pt
    ]
    {0.1*0.995^x};
    \addplot[
    domain=100:200,
    samples=500,
    color=orange,
    line width=1.05pt
    ]
    {0.1*0.995^100*0.96^(x-100)};
    \addlegendentry{Two-stage exponential decay}
    
    \end{axis}
    \end{tikzpicture}
  \caption{Comparison of exponential decay with $\lambda = 0.98$, step decay and two-stage exponential decay with  $\lambda_1 = 0.995, \lambda_2 = 0.96$. It is obvious that the reduced speed of exponential decay is more higher than step decay initially. } 
  \label{fig_exponential_decay_curve}
\end{figure}

SGD is one of the most popular optimizers for training DNNs. The so-called \textit{step decay learning rate scheduler} is usually used in SGD to change learning rates to specific values for different stages. For example, the step decay scheduler in \cite{ResNet_1} scheduled the learning rate as follows:
\begin{equation} \label{eq_step_decay}
lr = \left\{
\begin{array} {lc}
0.1 & 0 < e \leq 81 \\ 
0.01 & 81 < e \leq 122 \\ 
0.001 & 122 < e \leq 200
\end{array} \right.
\end{equation}
where $e$ refers to the index of the current epoch. Ideally, the step decay scheduler can achieve a good performance by carefully changing the learning rate and stage periods, but trial and error is needed to find an acceptable step decay scheduler. Another traditional scheduler is the \textit{exponential decay scheduler}. The exponential decay scheduler reduces the learning rate by a given factor at each iteration or epoch. The learning rate formula for the exponential decay scheduler is:
\begin{equation} \label{eq_exponential_decay}
lr_t = lr_0 \times \lambda^{t}
\end{equation}
where $lr_t$ refers to the learning rate at time $t$, $lr_0$ is the initial learning rate, and $\lambda \in [0,1]$ is a \textit{discount factor}. Figure~\ref{fig_exponential_decay_curve} illustrates the learning rate curves of step decay scheduler and exponential decay scheduler.

Momentum \cite{Momentum} is designed to accelerate DNN training. The momentum algorithm records the gradients in the past iterations, and combines them with the current gradient to decide where to move in this iteration. We need to decide the hyperparameter $ \beta \in [0,1] $ in the momentum algorithm to determine the contributions of past gradients to the current update. Nesterov Momentum, a variant of the momentum algorithm, is proposed by Nesterov \cite{Nesterov_theory}, and used to train DNN  by Sutskever \etal \cite{Nesterov_DNN}. The gradient when using Nesterov Momentum is evaluated after applying the current velocity. It can add a correction factor to the standard momentum algorithm. 

Adaptive learning rate methods usually adjust weights in a DNN based on a mechanism which requires users to determine some hyperparameters initially, but not for the overall design of the learning scheduler. There have been many different adaptive learning rate methods, including AdaGrad \cite{AdaGrad}, RMSProp \cite{RMSProp}, AdaDelta \cite{ADADELTA}, Adam \cite{Adam} and Adamax \cite{Adam}. These adaptive learning rate methods do not require hyperparameters fine-tuning to obtain proper learning rate value, this comes at a significant computing cost, while the final performance tends to be inferior to fine-tuned step decay. Shirish and Richard proposed a method where the learning rate is defined with Adam from the start, and by SGD towards the end \cite{SGD_Adam}, combining the benefits of a fast convergent rate and with the superior performance of step decay. These kinds of methods reduce the complexity of setting the learning rate scheduler, but the performances are worse than step decay in most cases.

SGDR \cite{SGDR}, CLR \cite{CLR} and ES-Learning \cite{ESLearning} are similar. They all used warm restart mechanisms to reset the learning rate every some epochs (or iterations) and demonstrated that these mechanisms improved the performances. In \cite{SGDR}, SGDR also proposed a \textit{cosine learning rate scheduler}, which follows the cosine wave from the maximum to the minimum and makes the change of learning rate smooth. Cosine scheduler schedules the learning rate as follow:

\begin{equation} \label{eq_cosine}
\begin{split} 
lr_t^{\textrm{cos}} &= lr_{\textrm{min}} +  \frac{lr_{\textrm{\textrm{max}}} - lr_{\textrm{min}}}{2}  \left( 1+ \cos \left( \frac{ \pi \times t}{T} \right) \right) 
\end{split}
\end{equation}

where $lr_{\textrm{min}}$ and $lr_{\textrm{max}}$ are the minimum and maximum learning rates respectively, $T$ is the total number of training epochs or iterations, and $t$ is the index of the current epoch or iteration. Figure~\ref{fig_HTD_curve} illustrates the cosine with $lr_{\textrm{min}}=0, lr_{\textrm{max}}=0.1$. In \cite{Snapshot, ConDenseNet, SGDR, DenseNet_Memory}, cosine learning rate scheduler outperformed step decay scheduler. However, the cosine learning rate scheduler is less flexible than step decay since only the maximum and minimum learning rates can be changed.

\section{Hyperbolic-Tangent Decay}
\label{sec_HTD}

This section proposes a new scheduling method, named \textit{hyperbolic-tangent decay} (\textit{HTD}). Subsection \ref{subsec_step_decay} first analyzes the performance of step decay with different settings, and Subsection \ref{subsec_exponential_decay} analyzes the performance of exponential decay and compares it with step decay. The analyses of Subsection \ref{subsec_step_decay} and Subsection \ref{subsec_exponential_decay} motivate the design of HTD described in Subsection \ref{htd_2}

\subsection{Step decay } 
\label{subsec_step_decay}

This subsection presents the performance analysis of step decay. We train Residual Network with 32 layers, denoted by ResNet-32, on CIFAR-10 with the following settings. The learning rate is 0.1 for the first S$_1$ epochs and 0.01 for the next S$_2$ epochs. For comparison, use different S$_1$ and S$_2$ such that S$_1$ + S$_2$ = 200. Then compare their averaged test error rates of 4 runs. 

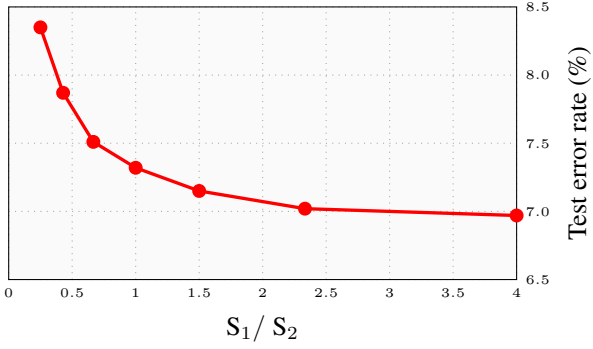
\begin{figure}[t]
	\centering
    \begin{tikzpicture}
        \begin{axis}
    [
        xlabel={S$_1 / $ S$_2$ },
		ylabel={Test error rate (\%)},
        yticklabel pos=right,
        yticklabel style = {font=\tiny},
        xticklabel style = {font=\tiny},
        xtick={0,0.5,...,4},
        ytick={6.0,6.5,...,8.5},
        xmin=0,
        xmax=4,
        ymin=6.5,
        ymax=8.5,
        width=\linewidth ,
        height=0.625\linewidth ,
        tick style={draw=none},
        y tick label style={
          /pgf/number format/.cd,
              fixed,
              fixed zerofill,
              precision=1,
          /tikz/.cd
        },
        grid,
        grid style={dotted,gray!70},
        yticklabel style = {font=\tiny},
        xticklabel style = {font=\tiny},
        legend style={font=\tiny,legend pos=north east,cells={anchor=west},fill=gray!4},
        axis background/.style={fill=gray!4}
    ]
    
		\addplot[
        color=red,
        mark=*,
        mark options={solid},
        line width=1.25pt
        ]
		coordinates {
        (40/160,8.35)
        (60/140,7.87)
        (80/120,7.51)
        (100/100,7.32)
        (120/80,7.15)
        (140/60,7.02)
        (160/40,6.97)
        };
      
    \end{axis}
    \end{tikzpicture}
  \caption{Test error rates of different ratios $\textrm{S}_1 / \textrm{S}_2$ with S$_1$ + S$_2$ = 200. The performances are significantly different for different ratios.}
  \label{step_decay_test_performance}
\end{figure}

The error rates of different ratios $\textrm{S}_1 / \textrm{S}_2$, shown in Figure~\ref{step_decay_test_performance}, indicate that the performances vary significantly as the ratio changes. In this experiment, the performance at the ratio 0.25 performs better than the one at 4 by reducing 1.38\% error rate. This result implies that it is important for a scheduler to choose the ratio flexibly for training.

\subsection{Exponential decay}
\label{subsec_exponential_decay}
This subsection analyzes the performance of exponential decay and compares it with step decay by training ResNet-32 on CIFAR-10 with 200 epochs and initial learning rate 0.1 as the previous subsection. For step decay, the learning rate is dropped by a factor of 0.1 at 81 and 122 epochs, like \cite{ResNet_1}. For exponential decay, the discount factor is 0.98, which makes the final learning rates of both exponential decay and step decay close. Each method runs 5 times and computes the averaged test error rates. Consequently, the averaged test error rates of exponential decay and step decay are $7.51\%$ and $7.18\%$ respectively in the experiments. Obviously, step decay performs better than exponential decay in this case. 

For further investigation, we try another experiment by proposing two-stage exponential decay as follows.
\begin{equation} \label{eq_two_stage_exponential_decay}
lr = \left\{
\begin{array} {lc}
lr_0\times \lambda_1^e &  e \leq 100 \\ 
lr_0 \times \lambda_1^{100} \times \lambda_2^{e-100} & e > 100
\end{array} \right.
\end{equation}
where $\lambda_1$ is 0.995 and $\lambda_2$ is 0.96 are two discount factors in this experiment, such that the final learning rates of step decay, exponential decay and two-stage exponential decay are close. Figure~\ref{fig_exponential_decay_curve} illustrates the learning rate curve of two-stage exponential decay. The figure shows that the learning rate curve of the first stage is quite flat while the decay speed for the second stage is faster. In this experiment, the averaged test error rates is 7.16\%, interestingly even lower than step decay. This result implies that the initial decay speed should not be too high.

\subsection{Hyperbolic-Tangent Decay (HTD)}
\label{htd_2}

From the empirical results in the previous subsections, this subsection proposes to use hyperbolic tangent functions for our learning rate scheduling as follow.
\begin{myfont}
\begin{equation} \label{eq_htd}
\begin{split}
lr_t^{\textrm{HTD}} &= lr_{\textrm{min}} + \frac{lr_{\textrm{max}} - lr_{\textrm{min}}}{2} \left( 1- \tanh \left( L + (U-L) \frac{t}{T} \right) \right) \\
&= lr_{\textrm{min}} + \frac{lr_{\textrm{max}} - lr_{\textrm{min}}}{2} \left( 1- \tanh \left( L \left(1-\frac{t}{T}\right) + U \frac{t}{T} \right) \right) 
\end{split}
\end{equation} 
\end{myfont} \\where $lr_{\textrm{max}}$ and $lr_{\textrm{min}}$ are the maximum and minimum learning rates respectively, $T$ is the total number of epochs (or iterations), $0 \leq t \leq T$ is the index of epoch (or iteration), $L$ and $U$ indicate the lower and upper bounds of the interval $[L,U]$ for the function  $\tanh x$. Figure \ref{fig_1-tanh} illustrates the function $1-\tanh x$  in the interval  $ [-6, 3]$. Let HTD($L, U, lr_{\textrm{max}}, lr_{\textrm{min}}$) denote the scheduler following the Formula \ref{eq_htd}. In this paper, $lr_{\textrm{min}}$ is set to 0 and $lr_{\textrm{max}}$ is set to the initial learning rate of step decay. For simplicity, let HTD($L,U$) denote the scheduler. In this paper, we only consider $L \leq 0$ and $U > 0$. Figure \ref{fig_HTD_curve} illustrates the two learning rate curves of HTD(-4,4) and HTD(-6,3).

\begin{figure*}[bh]
\begin{minipage}[t][0.6\columnwidth][t]{\columnwidth}
  \centering
  \begin{tikzpicture}
    \begin{axis}[
    xlabel={$x$},
    ylabel={$f(x)$},
    yticklabel style = {font=\tiny},
    xtick={-7,-5,...,7},
    ytick={0,0.5,...,2.5},
    width=0.9\columnwidth,
    height=0.54\columnwidth,
    xmin=-7,
    xmax=7,
    tick style={draw=none},
    y tick label style={
      /pgf/number format/.cd,
      fixed,
      fixed zerofill,
      precision=2,
      /tikz/.cd
    },
    grid,
    grid style={dotted,gray!70},
    yticklabel style = {font=\tiny},
    xticklabel style = {font=\tiny},
    legend style={font=\tiny,legend pos=north east,cells={anchor=west},fill=gray!4},
    axis background/.style={fill=gray!4}
    ] 
    
    \addplot [
    domain=-10:10, 
    smooth,
    color=black,
    line width=1.05pt
    ]
    {1 - tanh(x)};
    
    \addplot [
    domain=-6:3,
    smooth,
    color=red,
    line width=1.35pt
    ]
    {1 - tanh(x)};
    
	\addplot[red,mark=*]
    plot[error bars/.cd,
    y dir=minus,y fixed relative=1,
    x dir=minus,x fixed relative=1,
    error mark=none,
    error bar style={dotted}]
    coordinates
    {(-6,2)};
    
    \addplot[red,mark=*]
    plot[error bars/.cd,
    y dir=minus,y fixed relative=1,
    x dir=minus,x fixed relative=1,
    error mark=none,
    error bar style={dotted}]
    coordinates  {(3,0)};
 
   \end{axis}
    \end{tikzpicture}
    \caption{Function of $f(x)=1 - \tanh x$ with $L=-6$ and $U=3$.}
    \label{fig_1-tanh}
\end{minipage}
\hskip 0.8cm
\begin{minipage}[t][0.6\columnwidth][t]{\columnwidth}
  \begin{tikzpicture}
    \begin{axis}[
    xlabel={Epochs},
    ylabel={Learning rate},
    yticklabel style = {font=\tiny},
    xticklabel style = {font=\tiny},
    xtick={0,20,...,200},
    ytick={0,0.02,...,0.12},
    ymin=0,
    ymax=0.105,
    xmin=0,
    xmax=200,
    width=0.9\columnwidth,
    height=0.54\columnwidth,
    tick style={draw=none},
    y tick label style={
      /pgf/number format/.cd,
      fixed,
      fixed zerofill,
      precision=2,
      /tikz/.cd
    },
    grid,
    grid style={dotted,gray!70},
    yticklabel style = {font=\tiny},
    xticklabel style = {font=\tiny},
    legend style={font=\tiny,legend pos=south west,cells={anchor=west},fill=gray!4},
    axis background/.style={fill=gray!4}
    ] 
	\addplot [
    domain=0:200, 
    smooth,
    color=blue,
    line width=1.05pt
    ]
    {0.05*(1+cos(deg(pi*x/200)))};
	\addlegendentry{cosine}
    
    \addplot [
    domain=0:200, 
    smooth,
    color=orange,
    line width=1.05pt
    ]
    {0.05*(1-tanh(4*x/200 - 2))};
	\addlegendentry{HTD(-2,2)}
                
    \addplot [
    domain=0:200, 
    smooth,
    color=black,
    line width=1.05pt
    ]
    {0.05*(1-tanh(8*x/200 - 4))};
	\addlegendentry{HTD(-4,4)}
    
	\addplot [
    domain=0:200, 
    smooth,
    color=red,
    line width=1.05pt
    ]
    {0.05*(1-tanh(9*x/200 - 6))};
	\addlegendentry{HTD(-6,3)}

   \end{axis}
    \end{tikzpicture}
    \caption{Learning rate curves of HTD with different $L$ and $U$, and learning rate curve of cosine scheduler.}
    \label{fig_HTD_curve}
\end{minipage}
\end{figure*}

\paragraph{Close to two-stage exponential decay.} Next, we show that HTD is close to the two-stage exponential decay when $L\leq 0$ and $U > 0$. For simplicity, we use function $1-\tanh x$ to approximate. Note that $x=0$ is an inflection point of the function $1-\tanh x$.  We show that (a) the value of $1-\tanh x$ decreases slowly before the inflection point, like the first stage of two-stage exponential decay, and (b) the value of $1-\tanh x$ drops like exponential decay after the inflection point, like the second stage of two-stage exponential decay. Given displacement $\delta > 0$, we define the decreasing ratio of $1-\tanh x$ as:
\begin{myfont}
\begin{equation}
\begin{split}
r(x,\delta) := & \frac{1-\tanh (x+\delta)}{ 1-\tanh x} = \frac{2e^{-x-\delta}}{e^{x+\delta} + e^{-x-\delta}} \frac{e^{x} + e^{-x}}{2e^{-x}} \\
= & \frac{e^{-\delta}(e^x + e^{-x})}{e^{x+\delta}+e^{-x-\delta}} = \frac{e^{x-\delta} + e^{-x-\delta}}{e^{x+\delta} + e^{-x-\delta}} = \frac{e^{2x}+1}{e^{2x+2\delta}+1}
\end{split} 
\end{equation}
\end{myfont} \\for any position $x\in \mathbb{R}$. We observe the decreasing ratio for different position $x\in \mathbb{R}$. For (a), since $e^{2x+2\delta} \approx 0$ and $e^{2x} \approx 0$ when $x<0$ is sufficiently small, the decreasing ratio $r(x,\delta) \approx 1$ and hence decreasing speed is very slow, like the first stage of two-stage exponential decay. For (b), since $r(x,\delta)\approx  \frac{1}{e^{2\delta}}$ when $x>0$ is sufficiently large, the decreasing speed is close to exponential decay, like the second stage of two-stage exponential decay.

\paragraph{Compared to cosine.} The hyperparameters of cosine scheduler include $lr_{\textrm{max}}$ and $lr_{\textrm{min}}$. while the hyperparameters of HTD include $lr_{\textrm{max}}, lr_{\textrm{min}}, L$ and $U$. The additional hyperparameters of HTD are $L$ and $U$, which can determine the ratio of training time before and after the inflection point. These additional hyperparameters make HTD more flexible than cosine scheduler. We also find that HTD(-2,2) is close to the cosine scheduler, which is illustrated in Figure \ref{fig_HTD_curve}.

\begin{figure*}[hb]
\begin{minipage}[t][0.6\columnwidth][t]{\columnwidth}
  \centering
  \begin{tikzpicture}
    \begin{axis}[
    xlabel={Epochs},
    ylabel={Learning rate},
    yticklabel style = {font=\tiny},
    xticklabel style = {font=\tiny},
    xtick={0,20,...,200},
    ytick={0,0.02,...,0.12},
    ymin=0,
    ymax=0.105,
    xmin=0,
    xmax=200,
    width=0.9\columnwidth,
    height=0.54\columnwidth,
    tick style={draw=none},
    y tick label style={
      /pgf/number format/.cd,
      fixed,
      fixed zerofill,
      precision=2,
      /tikz/.cd
    },
    grid,
    grid style={dotted,gray!70},
    yticklabel style = {font=\tiny},
    xticklabel style = {font=\tiny},
    legend style={font=\tiny,legend pos=south west,cells={anchor=west},fill=gray!4},
    axis background/.style={fill=gray!4}
    ] 
    
    \addplot [
    domain=0:200, 
    smooth,
    color=black,
    line width=1.05pt
    ]
    {0.05*(1-tanh(4.5*x/200 - 1.5))};
	\addlegendentry{$R=0.5$}
    
    \addplot [
    domain=0:200, 
    smooth,
    color=blue,
    line width=1.05pt
    ]
    {0.05*(1-tanh(6*x/200 - 3))};
	\addlegendentry{$R=1$}
	
	\addplot [
    domain=0:200, 
    smooth,
    color=red,
    line width=1.05pt
    ]
    {0.05*(1-tanh(9*x/200 - 6))};
	\addlegendentry{$R=2$}

   \end{axis}
   \end{tikzpicture}

    \caption{Different $R$ with $U=3$. These curves have same final learning rates but different trends}
    \label{fig_R}
\end{minipage}
\hskip 0.8cm
\begin{minipage}[t][0.7\columnwidth][t]{\columnwidth}
 	\begin{tikzpicture}

 	\begin{axis}[
    xlabel={Epochs},
    ylabel={Learning rate},
    yticklabel style = {font=\tiny},
    xticklabel style = {font=\tiny},
    xtick={0,20,...,200},
    ytick={0,0.02,...,0.12},
    ymin=0,
    ymax=0.105,
    xmin=0,
    xmax=200,
    width=0.9\columnwidth,
    height=0.54\columnwidth,
    tick style={draw=none},
    y tick label style={
      /pgf/number format/.cd,
      fixed,
      fixed zerofill,
      precision=2,
      /tikz/.cd
    },
    grid,
    grid style={dotted,gray!70},
    yticklabel style = {font=\tiny},
    xticklabel style = {font=\tiny},
    legend style={font=\tiny,legend pos=south west,cells={anchor=west},fill=gray!4},
    axis background/.style={fill=gray!4}
    ] 

	\addplot [
    domain=0:200, 
    smooth,
    color=black,
    line width=1.05pt
    ]
    {0.05*(1-tanh(6*x/200 - 4))};
	\addlegendentry{$U=2$}
    
    \addplot [
    domain=0:200, 
    smooth,
    color=blue,
    line width=1.05pt
    ]
    {0.05*(1-tanh(9*x/200 - 6))};
	\addlegendentry{$U=3$}
	
	\addplot [
    domain=0:200, 
    smooth,
    color=red,
    line width=1.05pt
    ]
    {0.05*(1-tanh(12*x/200 - 8))};
	\addlegendentry{$U=4$}

   \end{axis}
    \end{tikzpicture}
    \caption{Different $U$ with $R=2$. These curves have similar trends but different final learning rates. The final learning rates of $U=2, U=3$ and $U=4$ are 0.0180, 0.0025 and 0.0003 respectively.}
    \label{fig_U}
\end{minipage}
\end{figure*}

\paragraph{Importance of $\textbf{\textit{L, U}}$ and their ratio.} In HTD($L,U$), the hyperparameters $U$ affects the final learning rate. For example, when $U=3$, the final learning rate is $lr_{\textrm{max}} \cdot (1-\tanh 3) \approx lr_{\textrm{max}} \cdot 0.005$. Besides, we can adjust the lower bound $L$ to change the ratio $R=|L|/|U|$, which is the ratio of training times before and after the inflection point. Like Subsection \ref{subsec_step_decay}, when the ratio $R$ is larger, the training time before the inflection point is longer. Figure \ref{fig_R} illustrates the learning rate curves with same  $U$ but different $R$. Figure \ref{fig_U} illustrates the learning rate curves with same $R$ but different $U$. 

\section{Experiments}
\label{experiments}

We empirically demonstrate effectiveness of HTD on several benchmark datasets and networks by comparing HTD with step decay scheduler and cosine scheduler.
\subsection{Datasets}

\paragraph{CIFAR.} Two CIFAR datasets \cite{CIFAR-10}, CIFAR-10 and CIFAR-100, consist of 60,000 color images with $32\times32$ pixels, 50,000 for training and 10,000 for testing.  CIFAR-10 images are classified into 10 classes with labels and CIFAR-100 images are 100 classes. For image preprocessing, we normalize the input data using the means and standard deviations based on \cite{WRN}. For data augmentation, we perform random crops from the image padded by 4 pixels on each side, filling missing pixels with reflections of the original image and horizontal flips with $50\%$ probability.

\paragraph{Fashion-MNIST.} Fashion-MNIST \cite{FASHION} is a MNIST-like \cite{MNIST} dataset, which consists of 70,000 gray-scale images with $28\times28$ pixels, classified into 10 categories each with 7,000 images. There are 60,000 and 10,000 images for training and testing respectively. For image preprocessing, we normalize the input data using the mean and standard deviation based on \cite{WRN}. For data augmentation, we also perform random crops from the image padded by 3 pixels on each side, filling missing pixels with reflections of the original image and horizontal flips with 50\% probability.

\paragraph{PAMAP2.} PAMAP2 \cite{PAMAP2} is a physical activity monitoring dataset, and consists of raw input recorded by sensor. It has 9 subjects, 8 males and 1 females with ages between 23 and 31, wearing 3 IMUs and a HR-monitor. We choose 12 (lie, sit, stand, iron, vacuum, ascend stairs, descend stairs, normal walk, nordic walk, cycle, run, and rope jump) from all 18 activities. For data preprocessing, we down sample data from 100Hz to 20Hz and choose all of data without data missing. $70\%$ of data are used to train, and $30\%$ of data are used to test.

\paragraph{ImageNet.} ILSVRC dataset \cite{IMAGENET} consists of 1000 classes and is split into three sets: 1.28 million training images, 50,000 validation images and 50,000 testing images. Each image is a $224 \times 224$ color image. Image preprocessing and data augmentation follow the settings of \cite{ResNet_1}. We adopt 1-crop and evaluate both top-1 and top-5 error rates in testing.

\subsection{Implementation}

\paragraph{Architectures.} We run experiments on several benchmarks including Residual Network (ResNet) \cite{ResNet_1, ResNet_2}, Wide Residual Network (Wide ResNet) \cite{WRN} and Densely Network (DenseNet) \cite{DenseNet}. Let ResNet-d denote the ResNet with depth d, WRN-d-k denote the Wide ResNet with depth d and width k, and DenseNet-BC-L-k denote the DenseNet with depth L and growth rate k. For DenseNet, BC represents the use of bottleneck mechanism with 50\% reduced rate. For PAMAP2, we train with a Bidirectional-LSTM (BLSTM) \cite{BLSTM} based on \cite{HAR}.

\paragraph{Optimizer.} All the networks are trained by SGD, using a Nesterov momentum \cite{Nesterov_DNN} of 0.9 and adopt the weight initialization introduced by \cite{he_normal_initial}. DenseNet is trained with a mini-batch size of 64, ResNet and Wide ResNet are trained with a mini-batch size of 128 except ImageNet, which is trained with a mini-batch size of 256. For ResNet, we use a weight decay of 0.0001, and start with a learning rate of 0.1, reduced by a factor of 0.1 at 81 and 122 epochs in step decay. For Wide ResNet, we use a weight decay of 0.0005 according to \cite{WRN}, and start with a learning rate of 0.1, reduced by a factor of 0.2 at 60, 120 and 160 epochs in step decay. For DenseNet, we use weight decay of 0.0001, and start with a initial learning rate 0.01, reduced by a factor of 0.1 at 50\% and 75\% of the total number of training epochs in step decay. ResNet and Wide ResNet are trained with 200 epochs, while DenseNet is trained with 300 epochs. 

\paragraph{Methods.} We compares three types of schedulers in following experiments: step decay, cosine scheduler (proposed by \cite{SGDR}) and HTD. Both cosine and HTD set the minimal learning rate to 0, and use the initial learning rate of step decay for the maximum learning rate. For HTD, we use the two versions HTD(-4,4) and HTD(-6,3) only.

\begin{table*}
\begin{center}
\begin{tabular}{|c|c|c|c|c|c}
\hline
Network  & Method & runs & \textbf{CIFAR-10} & \textbf{CIFAR-100} \\
\hline\hline
\multirow{4}{*}{ResNet-110} 
& step decay & med. of 5 & 6.03 & 27.49 \\
& cosine & med. of 5 & 5.91 & 27.00\\
& HTD(-4,4) & med. of 5 & 5.86 & 26.84 \\
& HTD(-6,3) & med. of 5 & {\color{blue}{\textbf{5.68}}} & {\color{blue}\textbf{26.78}} \\
\hline
\multirow{4}{*}{DenseNet-BC-100-12} 
& step decay & \cite{DenseNet} & $4.51^*$ & $22.27^*$ \\
& cosine     & med. of 5 & 4.51 & 22.59  \\
& HTD(-4,4) & med. of 5 & 4.45 & 22.47 \\
& HTD(-6,3) & med. of 5 & {\color{blue}{\textbf{4.43}}} & {\color{blue}{\textbf{22.17}}}\\ 
\hline
\multirow{4}{*}{WRN-28-10} 
& step decay & med. of 5 & 4.32 & 20.43   \\
& cosine & med. of 5 & 4.31 & 20.54 \\
& HTD(-4,4) & med. of 5 & 4.31 & 19.74 \\
& HTD(-6,3) & med. of 5 & {\color{blue}{\textbf{4.22}}} & {\color{blue}{\textbf{19.73}}}  \\
\hline
\multirow{4}{*}{DenseNet-BC-250-24} 
& step decay & \cite{DenseNet} & $3.62^*$ & $17.60^*$  \\
& cosine     & med. of 2 & {\color{blue}{\textbf{3.39}}} & 17.44  \\
& HTD(-4,4) & med. of 2 & 3.40 & 17.29 \\
& HTD(-6,3) & med. of 2 & 3.42 & {\color{blue}{\textbf{17.02}}}  \\
\hline
\end{tabular}
\end{center}
\caption{The error rates(\%) of CIFAR-10 and CIFAR-100. Best results are written in {\color{blue}{\textbf{blue}}}. The character * indicates results are directly obtained from the original paper.}
\label{result of CIFAR}
\end{table*}

\subsection{Experiments result}

\subsubsection{Different Networks for CIFAR}

For CIFAR-10 and CIFAR-100, this paper tests them on ResNet-110, WRN-28-10, DenseNet-BC-100-12 and DenseNet-BC-250-24. For DenseNet, we directly use the source code given in \cite{DenseNet,DenseNet_Memory}, so we do not test step decay again, that is, we use the results in \cite{DenseNet} directly. For each setting, run 5 times, and then compare the averaged test error rates for all settings. Table \ref{result of CIFAR} shows that HTD(-6,3) outperforms all the others except for DenseNet-BC-250-24 on CIFAR-10, where the error rate of HTD(-6,3) is slightly higher than cosine scheduler by 0.03\%. Table \ref{result of CIFAR} also shows that HTD(-6,3) performs better than HTD(-4,4), except for DenseNet-BC-250-24 on CIFAR-10. However, HTD(-4,4) still outperforms step decay and cosine except for WRN-28-10 and DenseNet-BC-250-24 on CIFAR-10.

From above, HTD(-6,3) improves the performance over step decay by lowering the error rates significantly, \eg, improving 0.35\% for ResNet-110 on CIFAR-10, 0.2\% for DenseNet-BC-250-24 on CIFAR-10, 0.7\% for ResNet-110 and WRN-28-10 on CIFAR-100, and 0.58\% for DenseNet-BC-250-24 on CIFAR-100. In addition, HTD(-6,3) outperforms cosine scheduler, by 0.42\% for DenseNet-BC-100-12 and DenseNet-BC-250-24 on CIFAR-100, 0.81\% for WRN-28-10 on CIFAR-100. 

\begin{table}
\begin{center}
\resizebox{0.9\columnwidth}{!}{%
\begin{tabular}{|c|c|c|c|c|}
\hline
Network & Method & runs & top-1 err. & top-5 err. \\
\hline\hline
\multirow{2}{*}{ResNet-18}
& step decay & \cite{fb_torch} & $30.43^*$ & $10.76^*$  \\
& HTD(-6,3) & med. of 2 & {\color{blue}{\textbf{29.84}}} & {\color{blue}{\textbf{10.49}}}  \\
\hline
\multirow{2}{*}{ResNet-34}
& step decay & \cite{fb_torch} & $26.73^*$ & $8.74^*$  \\
& HTD(-6,3) & med. of 1 & {\color{blue}{\textbf{26.25}}} & {\color{blue}{\textbf{8.43}}}  \\ 
\hline
\end{tabular}
}
\end{center}
\caption{The error rates (\%, 1-crop testing) on ImageNet. Best results are written in {\color{blue}{\textbf{blue}}}. The character * indicates results are directly obtained from the original paper.}
\label{table_ImageNet}
\end{table}

\begin{table*}
\begin{center}
\begin{tabular}{|c|c|c|c|}
\hline
datasets/Network & Method & runs & error rate \\
\hline\hline
\multirow{4}{*}{PAMAP2/BLSTM}
& step decay & med. of 5 & 11.88  \\
& cosine & med. of 5 & 11.94 \\
& HTD(-4,4) & med. of 5 & 11.79 \\
& HTD(-6,3) & med. of 5 & {\color{blue}{\textbf{11.72}}}  \\
\hline
\multirow{4}{*}{Fashion-MNIST/ResNet-110}
& step decay & med. of 5 & 4.99  \\
& cosine & med. of 5 & 4.90 \\
& HTD(-4,4) & med. of 5 & 4.92 \\
& HTD(-6,3) & med. of 5 & {\color{blue}{\textbf{4.84}}} \\
\hline
\end{tabular}
\end{center}
\caption{The error rates (\%) on PAMAP2 and Fashion-MNIST. Best results are written in {\color{blue}{\textbf{blue}}}.}
\label{table_other_dataset}
\end{table*}

\begin{figure*}[t]
\begin{minipage}[t][0.6\columnwidth][t]{\columnwidth}
	\centering
    \begin{tikzpicture}
        \begin{axis}
    [
        xlabel={Ratio $R$},
		ylabel={Test error rate (\%)},
        yticklabel pos=right,
        yticklabel style = {font=\tiny},
        xticklabel style = {font=\tiny},
        xtick={0,...,10},
        ytick={6.0,6.5,...,8.0},
        xmin=0,
        xmax=10,
        ymin=6,
        ymax=8,
        width=0.9\columnwidth,
        height=0.54\columnwidth,
        tick style={draw=none},
        y tick label style={
          /pgf/number format/.cd,
              fixed,
              fixed zerofill,
              precision=1,
          /tikz/.cd
        },
        grid,
        grid style={dotted,gray!70},
        yticklabel style = {font=\tiny},
        xticklabel style = {font=\tiny},
        legend style={font=\tiny,legend pos=north east,cells={anchor=west},fill=gray!4},
        axis background/.style={fill=gray!4}
    ]
    
		\addplot[
        color=red,
        mark=*,
        mark options={solid},
        line width=1.25pt
        ]
		coordinates {
		(0.5,7.16)
		(1,6.93)
		(2,6.72)
		(3,6.61)
		(5,6.56)
		(10,6.64)
        };
        \addlegendentry{trained by 200 epochs}
        
        \addplot[
        color=blue,
        mark=*,
        mark options={solid},
        line width=1.25pt
        ]
		coordinates {
		(0.5,6.64)
		(1,6.43)
		(2,6.14)
		(3,6.33)
		(5,6.14)
		(10,6.19)
        };
        \addlegendentry{trained by 400 epochs}
      
    \end{axis}
    \end{tikzpicture}
  \caption{Different ratio $R$ with same upper bound $U=3$ on CIFAR-10.}
  \label{fig_different_R_experiment_CIFAR10}
\end{minipage}
\hskip 0.8cm
\begin{minipage}[t][0.6\columnwidth][t]{\columnwidth}
	\centering
    \begin{tikzpicture}
        \begin{axis}
    [
        xlabel={Ratio $R$},
		ylabel={Test error rate (\%)},
        yticklabel pos=right,
        yticklabel style = {font=\tiny},
        xticklabel style = {font=\tiny},
        xtick={0,...,10},
        ytick={27,...,33},
        xmin=0,
        xmax=10,
        ymin=27,
        ymax=33,
        width=0.9\columnwidth,
        height=0.54\columnwidth,
        tick style={draw=none},
        y tick label style={
          /pgf/number format/.cd,
              fixed,
              fixed zerofill,
              precision=1,
          /tikz/.cd
        },
        grid,
        grid style={dotted,gray!70},
        yticklabel style = {font=\tiny},
        xticklabel style = {font=\tiny},
        legend style={font=\tiny,legend pos=north east,cells={anchor=west},fill=gray!4},
        axis background/.style={fill=gray!4}
    ]
    
		\addplot[
        color=red,
        mark=*,
        mark options={solid},
        line width=1.25pt
        ]
		coordinates {
		(0.5,31.61)
		(1,30.60)
		(2,29.80)
		(3,29.59)
		(5,29.50)
		(10,28.90)
        };
        \addlegendentry{trained by 200 epochs}
        
        \addplot[
        color=blue,
        mark=*,
        mark options={solid},
        line width=1.25pt
        ]
		coordinates {
		(0.5,30.60)
		(1,30.16)
		(2,29.39)
		(3,29.59)
		(5,29.46)
		(10,28.97)
        };
        \addlegendentry{trained by 400 epochs}
      
    \end{axis}
    \end{tikzpicture}
  \caption{Different ratio $R$ with same upper bound $U=3$ on CIFAR-100.}
  \label{fig_different_R_experiment_CIFAR100}
\end{minipage}
\end{figure*}

\subsubsection{Other Data Sets}

\paragraph{ImageNet.} We only evaluate the ResNet-18 and ResNet-34 by using the source code of [6] since it takes a large amount of time to train ImageNet. For the same reason, we run HTD(-6,3) with 2 runs for ResNet-18 and 1 run for ResNet-34. For step decay, we directly use the results of \cite{fb_torch}. Table \ref{table_ImageNet} shows that HTD(-6,3) improve the performance over step decay significantly, e.g., improving 0.58\% and 0.48\% top-1 error for ResNet-18 and ResNet-34 respectively, and 0.33\% and 0.31\% top-5 error for ResNet-18 and ResNet-34 respectively.

\paragraph{Fashion-MNIST and PAMAP2.} For Fashion-MNIST, this paper uses ResNet-110 like CIFAR datasets. For the dataset PAMAP2, this paper uses BLSTM to train. For each setting, we run 5 times and compare their averaged test error rate. Table \ref{table_other_dataset} shows that HTD(-6,3) outperforms others on both PAMAP2 and Fashion-MNIST datasets. Compared to step decay, HTD(-6,3) improves 0.16\% and 0.15\% for PAMAP2 and Fashion-MNIST respectively. Compared to cosine scheduler, HTD(-6,3) improves 0.22\% and 0.06\% for PAMAP2 and Fashion-MNIST respectively.

\subsubsection{The effect of ratio in HTD}
\label{subsubsection_effect_of_ratio}
 
To demonstrate the importance of adjusting the ratio $R$, we design experiments similar to Subsection \ref{subsec_step_decay}. We train ResNet-32 on CIFAR-10 and CIFAR-100 with 200 epochs and 400 epochs and fix $U=3$. For each setting, use the averaged test error rate of 4 runs. Figure \ref{fig_different_R_experiment_CIFAR10} shows the performance on CIFAR-10, while Figure \ref{fig_different_R_experiment_CIFAR100} shows the performance on CIFAR-100.

Interestingly, the performance of CIFAR-10 improves significantly when the total epoch number doubles, i.e., $T=400$, while the performance of CIFAR-100 does not improve for $T=400$ when the ratio $R$ larger than 3. This implies that $T=200$ is not sufficient for CIFAR-10, but sufficient for CIFAR-100. For CIFAR-10, the best performance is $R=5$ for 200 epochs and $R=2$ for 400 epochs. For CIFAR-100, the best performance is $R=10$ for both 200 and 400 epochs. This implies that the best ratios are different for different datasets and settings of hyperparameters. 

\section{Conclusion}
\label{conclusion}
This paper proposes a new learning rate scheduler, HTD, on SGD. Our experiments show that HTD is superior to the step decay and cosine scheduler in the following aspects. Compared to step decay, HTD outperforms it in all experiments, and has less hyperparameters to tune. Compared to cosine scheduler, HTD has better performance in nearly all cases, and is more flexible to achieve better performance. Although different hyperparameters of the HTD have different performances, the experiments show that HTD(-6,3) is a good choice in most cases. Thus, HTD serves as one alternative for training a model by the SGD, and HTD(-6,3) is recommended as a default learning rate scheduler.

More researches on HTD are worthy investigating in the future. First, it is still an open problem to determine the best hyperparameters, such as $L$ and $U$, for many different datasets or networks. For example, Subsubsection \ref{subsubsection_effect_of_ratio} shows that $L=-15, U=3$ ($R=5$) and $L=-30, U=3$ ($R=10$) performs best on CIFAR-10 and CIFAR-100 respectively. This implies that the best $L$ and $U$ are different for different datasets. Hence, it is still likely to explore better results. Second, this paper has not yet tried some hybrid mechanisms for HTD. For example, incorporate HTD into the restart mechanisms like \cite{ESLearning,SGDR,CLR}. With restart mechanism, HTD is also able to use the snapshot like \cite{Snapshot} to improve the performance and avoid over-fitting.

{\small
\bibliographystyle{ieee}
\bibliography{thesis}
}

\end{document}